\documentclass{article}
\usepackage[utf8]{inputenc}
\usepackage{authblk}
\usepackage{setspace}
\usepackage[margin=1.25in]{geometry}
\usepackage{graphicx}
\graphicspath{ {./figures/} }
\usepackage{subcaption}
\usepackage{amsmath,amssymb,amsfonts}
\usepackage{dsfont}
\usepackage{threeparttable}
\usepackage{wrapfig}

\usepackage[style=ieee, 
citestyle=numeric-comp,
backend=bibtex,
sorting=none]{biblatex}
\title{MVCNet: Multiview Contrastive Network for Unsupervised Representation Learning for 3D CT Lesions}

\author[1,2]{Penghua Zhai}
\author[1,2]{Huaiwei Cong}
\author[3]{Gangming Zhao}
\author[4]{Chaowei Fang}
\author[1,2*]{Jinpeng Li}
\author[1, 2]{Ting Cai}
\author[5]{Huiguang He}

\affil[1]{Center for Pattern Recognition and Intelligent Medicine, HwaMei Hospital, University of Chinese Academy of Sciences, Ningbo 315010, China}
\affil[2]{Ningbo Institute of Life and Health Industry, University of Chinese Academy of Sciences, Ningbo 315010, China}
\affil[3]{Department of Computer Science, The University of Hong Kong, Hong Kong, China}
\affil[4]{School of Artificial Intelligence, Xidian University, Xi'an 710071, China}
\affil[5]{Institute of Automation, Chinese Academy of Sciences, No. 95 Zhongguancun East Road, Haidian District, Beijing, China}
\affil[*]{Corresponding author. Email: lijinpeng@ucas.ac.cn}
\date{}

\begin{document}

\maketitle

%%%%%% Abstract %%%%%%
\begin{abstract}
\emph{Objective and Impact Statement}. With the renaissance of deep learning, automatic diagnostic systems for computed tomography (CT) have achieved many successful applications. However, they are mostly attributed to careful expert annotations, which are often scarce in practice. This drives our interest to the unsupervised representation learning. \emph{Introduction}. Recent studies have shown that self-supervised learning is an effective approach for learning representations, but most of them rely on the empirical design of transformations and pretext tasks. \emph{Methods}. To avoid the subjectivity associated with these methods, we propose the MVCNet, a novel unsupervised three dimensional (3D) representation learning method working in a transformation-free manner. We view each 3D lesion from different orientations to collect multiple two dimensional (2D) views. Then, an embedding function is learned by minimizing a contrastive loss so that the 2D views of the same 3D lesion are aggregated, and the 2D views of different lesions are separated. We evaluate the representations by training a simple classification head upon the embedding layer. \emph{Results}. Experimental results show that MVCNet achieves state-of-the-art accuracies on the LIDC-IDRI (89.55\%), LNDb (77.69\%) and TianChi (79.96\%) datasets for \emph{unsupervised representation learning}. When fine-tuned on 10\% of the labeled data, the accuracies are comparable to the supervised learning model (89.46\% vs. 85.03\%, 73.85\% vs. 73.44\%, 83.56\% vs. 83.34\% on the three datasets, respectively). \emph{Conclusion}. Results indicate the superiority of MVCNet in \emph{learning representations with limited annotations}.
% \begin{itemize}
%     \item Objective: An opening sentence that states the objective of the research
%     \item Impact Statement: Brief description about the novelty and impact of the research
%     \item Introduction: Enough background information to give context to the study
%     \item Methods: A brief statement of the primary methods used by the study
%     \item Results: A brief statement of primary results
%     \item Conclusion: A short concluding sentence of the main take-home point(s) of the study
% \end{itemize}
\end{abstract}

%%%%%% Main Text %%%%%%

\section{Introduction}
% Your manuscript should contain all of the numbered sections specified in this template: Introduction, Results, Discussion, Materials and Methods.

% The manuscript should start with a brief introduction that lays out the problem addressed by the research and describes the paper’s importance. The scientific question being investigated should be described in detail. The introduction should provide sufficient background information to make the article understandable to readers in other disciplines and provide enough context to ensure that the implications of the experimental findings are clear.

Computed Tomography (CT) is one of the most widely used examinations in clinical applications. CT scans contain hundreds of slices, making it time-consuming for the physicians to browse and analyze them layer-by-layer. In addition, the image interpretation may vary in physicians, leading to ambiguity in decision-making. With the development of deep learning, computer-aided diagnosis (CAD) systems have greatly improved the efficiency and diagnostic accuracy of physicians. Currently, building a CAD system usually requires a large amount of annotated data. However, it is laborious and time-consuming for physicians to annotate CT scans. It is also difficult to aggregate data from different institutes due to the data island effect. Therefore, CAD systems are still confined by the lack of annotated volumetric data.

To cope with the lack of annotated data, researchers have attempted to exploit useful information from the unlabeled data with unsupervised learning algorithms \cite{dike2018unsupervised, pouyanfar2018survey, chellappa2021advances}. Recently, self-supervised learning (SSL), a new unsupervised learning paradigm, has attracted increasing attention due to its excellent representation learning ability \cite{noroozi2016unsupervised, caron2018deep, chen2020simple, he2020momentum, grill2020bootstrap, jing2020self, lee2020predicting, liu2021self}. As shown in Table S1, we compare several typical SSL in terms of three aspects: data type, transformation and pretext task. SSL can be divided into two categories: \emph{pretext task} methods \cite{gidaris2018unsupervised, zhu2020rubik, chen2019self, doersch2015unsupervised} and \emph{transformation-based} methods \cite{chen2020simple, he2020momentum, grill2020bootstrap, chen2021exploring}. The former aims to learn representations by the pseudo-labels, such as solving the Jigsaw puzzles \cite{noroozi2016unsupervised} and predicting spatial patches \cite{doersch2015unsupervised}. The latter typically use contrastive losses to endow the model with the invariance to transformations such as rotation, painting and Coloring. However, the pretext tasks and transformations are designed empirically, and different transformations or pretext tasks have different optimal configurations according to downstream tasks \cite{xiao2020should}.

Due to the complexity of medical imaging, it is important to consider expert knowledge when designing pretext tasks and transformations \cite{zhou2019models, zhu2020rubik, shen2017deep, li2020self}. For example, \cite{zhu2020rubik} learned representations by solving a Rubik's Cube task. \cite{zhou2019models} designed a series of transformations and learn representations by reconstructing the original image. Nevertheless, the transformations are subject to the human experience \cite{zhou2019models, xiao2020should}. Developing a new SSL paradigm for learning representations in 3D CT scans without transformations and pretext tasks is a challenging yet important task.
% Researchers usually take some prior knowledge into account when implementing contrast learning based on medical imaging \cite{}. Bai et al. \cite{} proposed a new method for learning features of cardiac MR image segmentation networks by predicting anatomical locations in a self-supervised manner, which is better or comparable than the scratch-trained U-net. Zhu et al. \cite{} proposed a novel self-supervised method for 3D medical data, named Rubik's Cube+. The Rubik's Cube+ framework forces the network to learn translational and rotational invariant features from the original 3D medical data via a Rubik's cube pretext task, while tolerating noise in the data.

\begin{figure}[h]
\centerline{\includegraphics[width=\linewidth]{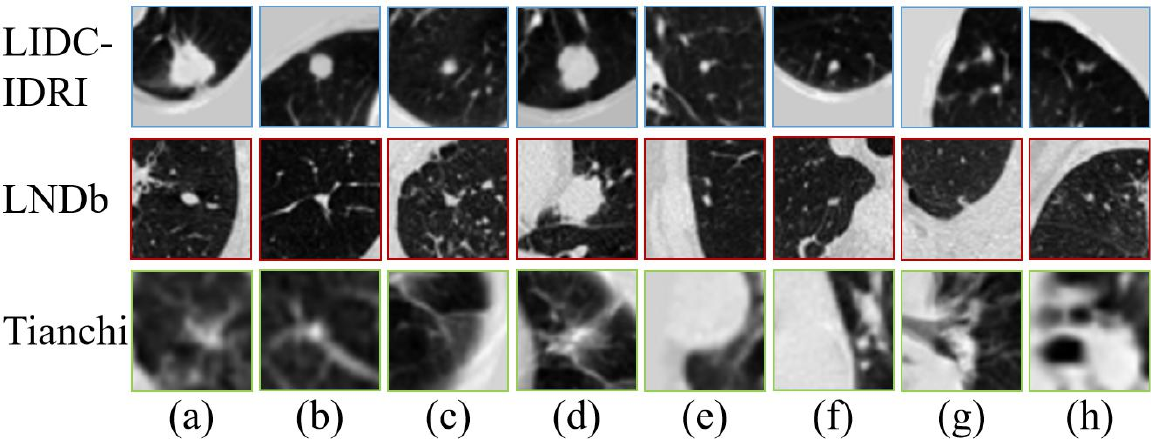}}
\caption{Examples of lesions in the LIDC-IDRI, LNDb, and TianChi datasets. For LIDC-IDRI and LNDb, (a)-(d) are the malignant nodules and (e)-(h) are the benign nodules. For TianChi, (a)-(b), (c)-(d), (e)-(f) and (g)-(h) are the lung nodules, streak shadow, arteriosclerosis (calcification) and lymph node calcification.}
\label{mvcnet_lesion}
\end{figure}

To deal with these challenges, we propose a novel multi-view contrastive network (MVCNet) for learning 3D CT representations. Different from traditional transformations, we extract nine views of each lesion from different orientations. This method is prevalent in recent years \cite{setio2016pulmonary, xie2018knowledge, xie2019semi, su15mvcnn, li2018survey, zhao2017multi}. We regard the views of the same lesion as \emph{positive pairs}, and the views of different lesions as \emph{negative pairs}. Inspired by the procedures of previous studies \cite{tian2020contrastive, chen2020simple}, we learn representations by minimizing a contrastive loss, resulting in an embedding space with the property of \emph{within-lesion compactness} and \emph{between-lesion separability}. We validate the effectiveness of our method on three lung CT datasets. Experimental results show that the proposed MVCNet outperforms state-of-the-art SSL methods.

The contribution of this paper is threefold. 1) We propose a multi-view contrastive learning framework for unsupervised representation learning on lung 3D CT data. 2) We exploit a transformation-free method for SSL, where the transformations are replaced by 2D views of 3D lesions from different orientations. To the best of our knowledge, this is the first work to implement 3D SSL without transformations and explicit pretext tasks in the medical image analysis field. 3) Extensive experiments on three open CT datasets (LIDC-IDRI, LNDb and TianChi) demonstrate the superiority of MVCNet over existing state-of-the-art SSL methods. Moreover, MVCNet obtains comparable results to the 3D supervised methods, indicting the gap between supervised learning and unsupervised learning has been largely bridged in our task. Figure. \ref{mvcnet_lesion} shows some lesions from the three datasets.

\section{Results}

\renewcommand{\arraystretch}{1.3}
\begin{table*}[h]
	\centering
	%	\fontsize{13}{18}
	\setlength{\tabcolsep}{1pt}
	\begin{threeparttable}
		\caption{Comparison of MVCNet with state-of-the-art SSL methods on the LIDC-IDRI}
		\label{table_linear_classification}
		\begin{tabular}{l c c c c c c}
			\hline\hline
			\multicolumn{1}{l}{\textbf{Methods}} & \textbf{AUC ($ \% $)} & \textbf{Sensitivity ($ \% $)} & \textbf{Specificity ($ \% $)} & \textbf{Accuracy ($ \% $)} & \textbf{Precision ($ \% $)} \\
			\hline\hline
			\multicolumn{4}{l}{Natural Image Augmentation} \\
			\hline
			Context \cite{doersch2015unsupervised} & $ 64.93 \pm 1.60 $ & $ 56.64 \pm 0.71 $ & $ 57.18 \pm 0.79 $ & $ 56.88 \pm 0.97 $ & $ 65.90 \pm 3.64 $ \\
			RotNet \cite{gidaris2018unsupervised} & $ 64.96 \pm 1.14 $ & $ 55.12 \pm 9.93 $ & $ 58.72 \pm 9.76 $ & $ 56.61 \pm 2.21 $ & $ 68.05 \pm 5.58 $ \\
			MoCo \cite{he2020momentum} & $ 71.07 \pm 0.11 $ & $ 70.26 \pm 0.46 $ & $ 71.58 \pm 0.45 $ & $ 71.29 \pm 0.13 $ & $ 69.21 \pm 0.33 $ \\
			MoCo V2 \cite{chen2020improved} & $ 71.88 \pm 0.33 $ & $ 70.08 \pm 0.50 $ & $ 73.27 \pm 0.70 $ & $ 71.83 \pm 0.25 $ & $ 70.20 \pm 0.50 $ \\
			SimCLR \cite{chen2020simple} & $ 78.58 \pm 0.52 $ & $ 73.80 \pm 1.55 $ & $ 83.36 \pm 0.66 $ & $ 79.04 \pm 0.50 $ & $ 79.65 \pm 0.84 $ \\
% 			SimCLR (1-9) & $ 78.95 \pm 0.67 $ & $ 74.28 \pm 1.83 $ & $ 83.62 \pm 0.89 $ & $ 79.06 \pm 0.54 $ & $ 79.82 \pm 0.43 $ & $ 76.38 \pm 0.89 $ \\
			BYOL \cite{grill2020bootstrap} & $ 78.35 \pm 0.22 $ & $ 60.53 \pm 0.66 $ & $ 71.18 \pm 0.64 $ & $ 65.19 \pm 0.51 $ & $ 72.99 \pm 0.52 $ \\
			SimSiam \cite{chen2021exploring} & $ 76.39 \pm 0.93 $ & $ 66.32 \pm 1.79 $ & $ 71.71 \pm 1.77 $ & $ 69.11 \pm 1.12 $ & $ 69.23 \pm 1.45 $ \\
			\hline\hline
			\multicolumn{4}{l}{Medical Image Augmentation} \\
			\hline
			MoCo \cite{he2020momentum} & $ 73.27 \pm 0.55 $ & $ 71.50 \pm 0.96 $ & $ 74.91 \pm 1.31 $ & $ 74.15 \pm 0.53 $ & $ 72.49 \pm 0.85 $ \\
			MoCo V2 \cite{chen2020improved} & $ 78.45 \pm 0.85 $ & $ 78.48 \pm 1.96 $ & $ 77.66 \pm 1.05 $ & $ 78.69 \pm 0.55 $ & $ 76.50 \pm 1.59 $ \\
			SimCLR \cite{chen2020simple} & $ 79.98 \pm 0.50 $ & $ 70.32 \pm 0.81 $ & $ \textbf{89.64} \pm 1.02 $ & $ 80.31 \pm 0.71 $ & $ 84.88 \pm 1.70 $ \\
			BYOL \cite{grill2020bootstrap} & $ 70.49 \pm 0.18 $ & $ 57.06 \pm 1.05 $ & $ 76.96 \pm 1.17 $ & $ 65.77 \pm 0.94 $ & $ 76.12 \pm 1.14 $ \\
			SimSiam \cite{chen2021exploring} & $ 77.10 \pm 1.12 $ & $ 66.79 \pm 1.24 $ & $ 71.11 \pm 0.79 $ & $ 70.32 \pm 1.69 $ & $ 69.23 \pm 1.34 $ \\
            Models Genesis \cite{zhou2019models} & $ 76.29 \pm 2.33 $ & $ 61.17 \pm 0.92 $ & $ 74.33 \pm 1.68 $ & $ 65.83 \pm 1.56 $ & $ 77.90 \pm 0.91 $ \\
			Rubik’s Cube+ \cite{zhu2020rubik} & $ 82.07 \pm 0.44 $ & $ 78.00 \pm 0.92 $ & $ 83.80 \pm 0.77 $ & $ 81.21 \pm 0.16 $ & $ 81.17 \pm 0.56 $ \\
			Restoration~ \cite{chen2019self} & $ 85.60 \pm 0.31 $ & $ 73.68 \pm 1.85 $ & $ 81.03 \pm 1.04 $ & $ 78.75 \pm 0.84 $ & $ 72.34 \pm 1.17 $ \\
			\hline\hline
			Ours \\
			\hline
			\textbf{MVCNet} & $ \textbf{88.74} \pm 0.23 $ & $ \textbf{85.92} \pm 0.87 $ & $ 88.51 \pm 0.81 $ & $ \textbf{89.55} \pm 0.42 $ & $ \textbf{88.87} \pm 0.67 $ \\
% 			\textbf{MVCNet}-Shared-9 & $ 86.14 \pm 0.41 $ & $ 82.16 \pm 0.55 $ & $ 88.87 \pm 0.35 $ & $ 86.57 \pm 0.34 $ & $ 88.01 \pm 0.21 $ & $ 84.55 \pm 0.40 $ \\
% 			\textbf{MVCNet}-Shared-8 & $ 86.17 \pm 0.46 $ & $ 82.02 \pm 0.88 $ & $ 89.24 \pm 1.34 $ & $ 86.67 \pm 0.36 $ & $ 88.69 \pm 1.25 $ & $ 84.63 \pm 0.45 $ \\
% 			MVCNet (fine-tune) & $ 91.79 \pm 0.34 $ & $ 88.28 \pm 1.23 $ & $ 91.18 \pm 1.10 $ & $ 89.52 \pm 0.34 $ & $ 89.24 \pm 1.49 $ & $ 90.06 \pm 0.50 $ \\
			\hline\hline
		\end{tabular}
	\end{threeparttable}
\end{table*}

\subsection{Linear Evaluation}
We evaluate the representations by training a linear classifier on top of the frozen representation, which is a common practice in previous studies \cite{tian2020contrastive, chen2020simple}. Three sets of different experiments on LIDC-IDRI are listed in Table \ref{table_linear_classification}, including SSL based on natural image augmentations (NIA), SSL based on medical image augmentations (MIA) and the proposed MVCNet using nine views. Considering the limitation of space, the results on LNDb and TianChi are listed in Table S3 and Table S4. We used the transformations in the SimCLR as the basic natural image augmentation and the augmentation in the Models genesis as the medical image augmentation (see Table S2). The Models genesis, Rubik's Cube+, and Restoration are originally developed for medical imaging, and the others are developed based on natural images. By comparing the metrics, we harvest the following observations:
% Considering the limitation of space, the results on LNDb and TianChi are listed in Appendix Table S7 \ref{table_linear_classification_lndb_appe} and Table S6 \ref{table_linear_classification_tianchi_appe}.
% We used the transformations in the SimCLR as the basic natural image augmentation and the augmentation in the Models genesis as the medical image augmentation (see Table \ref{table_augmentation}).

First, the performance of the methods based on medical image augmentation is higher than the method based on natural image augmentation when applying them to CT scans. For SimCLR and MoCo V2, we get a 1\% and 7\% improvement separately when changing the augmentations from natural image augmentations to medical image augmentations. Rubik's Cube+ obtains the highest accuracy among all the methods except the proposed MVCNet.

% most of the existing sota contrastive learning methods are based on transformations. 
Second, for the proposed MVCNet, we use the different views decomposed from lesion volumes instead of transformations to learn representations for 3D lesions. As is shown in Table \ref{table_linear_classification}, we obtain an accuracy of 89.55\% using nine views. Compared with SimCLR (NIA), SimCLR (MIA) and Rubik's Cube+, MVCNet achieves 10\%, 9\%, and 8\% improvement, respectively. The results show that the proposed MVCNet has superior performance comparing with previous state-of-the-arts in SSL.

\renewcommand{\arraystretch}{1.3}
\begin{table*}[h]
	\centering
	%	\fontsize{13}{18}
	\setlength{\tabcolsep}{1pt}
	\begin{threeparttable}
		\caption{Performance comparison of MVCNet using 10\% data for fine-tuning on LIDC-IDRI and LNDb}
		\label{table_fine-tune}
		\begin{tabular}{l c c c c c}
			\hline\hline
			\textbf{Methods} & \textbf{AUC ($ \% $)} & \textbf{Sensitivity ($ \% $)} & \textbf{Specificity ($ \% $)} & \textbf{Accuracy ($ \% $)} & \textbf{Precision ($ \% $)} \\
			\hline\hline
			\multicolumn{4}{l}{LNDb} \\
			\hline
			SimCLR (NIA) \cite{chen2020simple} & $ 73.62 \pm 0.35 $ & $ 71.62 \pm 0.83 $ & $ \textbf{75.63} \pm 1.13 $ & $ 72.97 \pm 0.99 $ & $ \textbf{87.43} \pm 0.67 $ \\
			SimCLR (MIA) \cite{chen2020simple} & $ 73.85 \pm 0.92 $ & $ 76.37 \pm 1.23 $ & $ 71.33 \pm 1.32 $ & $ 78.34 \pm 0.91 $ & $ 81.48 \pm 0.78 $ \\
			AlexNet \cite{krizhevsky2012imagenet} & $ 70.65 \pm 0.81 $ & $ 76.09 \pm 0.69 $ & $ 66.67 \pm 0.62 $ & $ 73.44 \pm 0.21 $ & $ 85.37 \pm 0.74 $ \\
			AlexNet (3D) \cite{krizhevsky2012imagenet} & $ \textbf{77.43} \pm 0.85 $ & $ \textbf{89.48} \pm 1.03 $ & $ 69.23 \pm 0.83 $ & $ \textbf{80.25} \pm 0.53 $ & $ 80.95 \pm 0.77 $ \\
% 			\hline
            % \textbf{MVCNet} (LE) & $ \textbf{77.49} \pm 0.48 $ & $ 81.33 \pm 0.92 $ & $ 71.74 \pm 0.98 $ & $ 77.69 \pm 0.36 $ & $ 82.43 \pm 0.71 $ & $ 81.88 \pm 0.55 $ \\
			\textbf{MVCNet} & $ 70.12 \pm 1.07 $ & $ 84.73 \pm 0.29 $ & $ 53.27 \pm 0.55 $ & $ 73.85 \pm 0.41 $ & $ 78.05 \pm 0.80 $ \\
			\hline\hline
			\multicolumn{4}{l}{LIDC-IDRI} \\
			\hline
			SimCLR (NIA) \cite{chen2020simple} & $ 85.91 \pm 0.74 $ & $ 83.42 \pm 1.23 $ & $ 88.41 \pm 0.78 $ & $ 84.90 \pm 0.73 $ & $ 87.35 \pm 0.42 $ \\
			SimCLR (MIA) \cite{chen2020simple} & $ 85.47 \pm 0.17 $ & $ 85.21 \pm 0.86 $ & $ 85.74 \pm 0.80 $ & $ 85.46 \pm 0.22 $ & $ 84.48 \pm 0.60 $ \\
			AlexNet \cite{krizhevsky2012imagenet} & $ 90.90 \pm 0.47 $ & $ 82.42 \pm 1.55 $ & $ 87.05 \pm 1.05 $ & $ 85.03 \pm 0.50 $ & $ 85.52 \pm 0.95 $ \\
			AlexNet (3D) \cite{krizhevsky2012imagenet} & $ 78.88 \pm 0.31 $ & $ 72.46 \pm 0.95 $ & $ 79.32 \pm 0.12 $ & $ 76.60 \pm 0.42 $ & $ 75.56 \pm 0.65 $ \\
% 			\hline
			\textbf{MVCNet} & $ \textbf{91.79} \pm 0.34 $ & $ \textbf{88.28} \pm 1.23 $ & $ \textbf{91.18} \pm 1.10 $ & $ \textbf{89.46} \pm 0.54 $ & $ \textbf{89.24} \pm 1.49 $ \\
			\hline\hline
		\end{tabular}
		NIA: natural image augmentation. MIA: medical image augmentation. LE: linear evaluation. AlexNet and AlexNet (3D) are trained on the annotated data.
	\end{threeparttable}
\end{table*}

\subsection{Performance with Fine-tuning}
\begin{figure}[!t]
\centerline{\includegraphics[width=\linewidth]{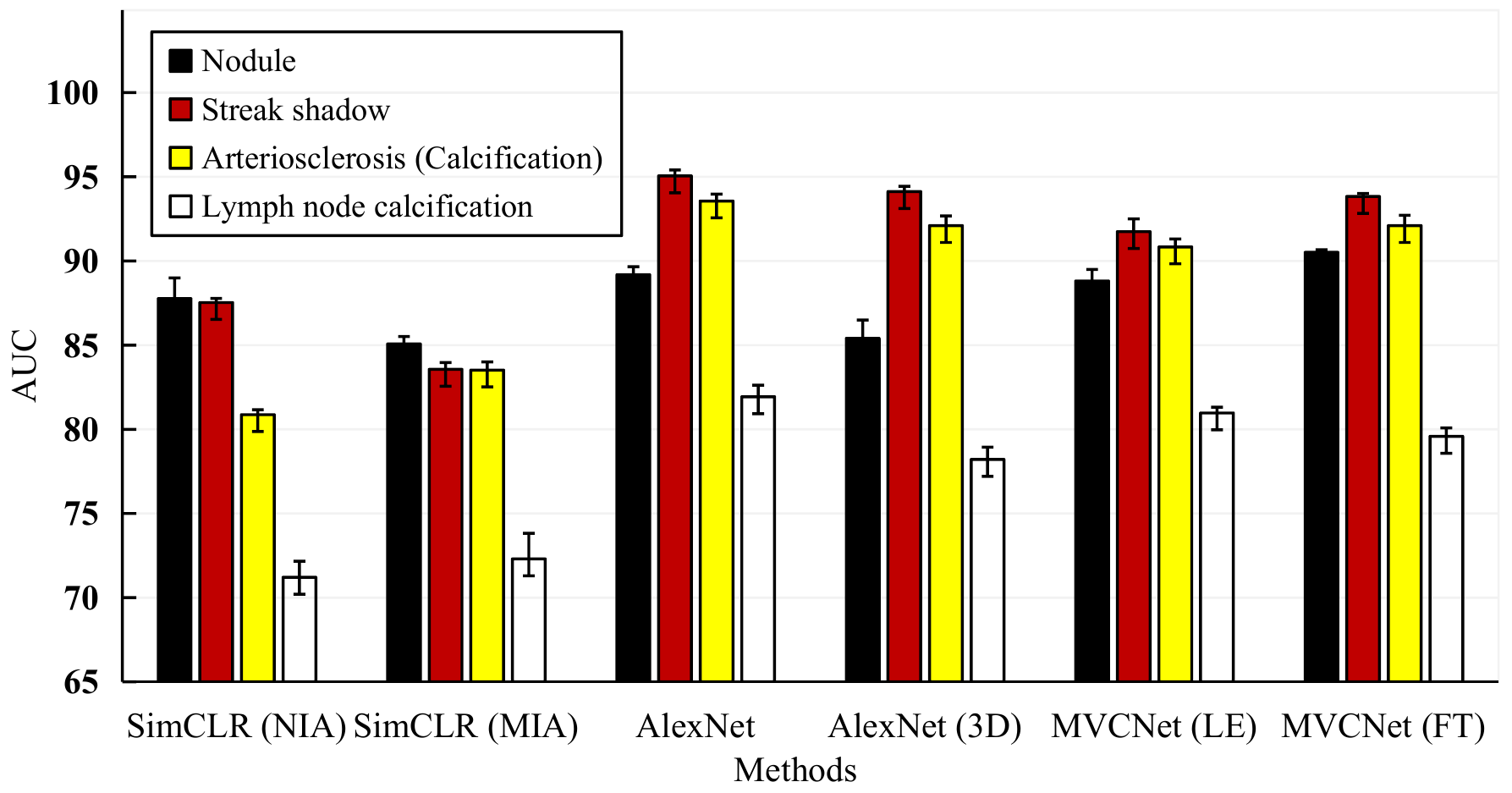}}
\caption{Results of fine-tuning with 10\% data on TianChi dataset. The vertical axis represents the diagnosis results evaluated by AUC. The horizontal axis represents different methods. \textbf{NIA}: natural image augmentation, \textbf{MIA}: medical image augmentation. \textbf{LE}: linear evaluation, \textbf{FT}: fine-tuning. Different colors represent different diseases (black: Nodules; red: Streak shadows; yellow: Arteriosclerosis (Calcification); white: Lymph node calcification).}
\label{tianchi_binary_cls}
\end{figure}

To further evaluate the effectiveness of the MVCNet with limited data, we fine-tune the model with 10\% annotated samples in datasets. This protocol has been used in some previous studies \cite {chen2020simple}. In Table \ref{table_fine-tune}, Figure \ref{tianchi_binary_cls} and Table \ref{table_tianchi_multi_classification}, we fine-tune SimCLR with 10\% datasets and compare the results with the fully-supervised methods using all the samples.

As shown in Table \ref{table_fine-tune}, the accuracies are 89.46\% and 73.85\% on LIDC-IDRI and LNDb, respectively. For LIDC-IDRI, MVCNet shows advantages over SimCLR and AlexNet with large margins (4\% improvement, respectively). For LNDb, MVCNet is better than AlexNet while but inferior to the 3D AlexNet. As is shown in Figure \ref{tianchi_binary_cls}, we diagnose each of the four diseases and use AUC as the metric due to the class imbalance. Figure \ref{tianchi_binary_cls} demonstrates that the performance of MVCNet is better than SimCLR and is comparable to the fully-supervised AlexNet. We also conduct multi-disease diagnosis experiments on the TianChi dataset, as is shown in Table \ref{table_tianchi_multi_classification}. The diagnostic accuracy of MVCNet is slightly higher (0.22\%) than that of AlexNet, and it is higher (3\%) than that of SimCLR. Overall, the MVCNet fine-tuned with 10\% datasets is better than SimCLR and is comparable to the fully-supervised AlexNet.

\renewcommand{\arraystretch}{1.3}
\begin{table}[h]
	\centering
	%	\fontsize{13}{18}
	\setlength{\tabcolsep}{40pt}
	\begin{threeparttable}
		\caption{Performance comparison of MVCNet using 10\% data for fine-tuning on TianChi for multi-diseases diagnosis.}
		\label{table_tianchi_multi_classification}
		\begin{tabular}{l c}
			\hline\hline
			\textbf{Methods} & \textbf{Accuracy ($ \% $)} \\
			\hline
			SimCLR (NIA) \cite{chen2020simple} & $ 79.08 \pm 0.56 $ \\
			SimCLR (MIA) \cite{chen2020simple} & $ 80.72 \pm 0.93 $ \\
			AlexNet \cite{krizhevsky2012imagenet} & $ 83.34 \pm 0.33 $ \\
			AlexNet (3D) \cite{krizhevsky2012imagenet} & $ 78.13 \pm 0.75 $ \\
% 			\hline
% 			\textbf{MVCNet} (LE) & $ 79.96 \pm 0.84 $ \\
			\textbf{MVCNet} & $ \textbf{83.56} \pm 0.68 $ \\
			\hline\hline
		\end{tabular}
		NIA: natural image augmentation. MIA: medical image augmentation. LE: linear evaluation. FT: fine-tuning.
	\end{threeparttable}
\end{table}

\begin{figure}[h]
\centerline{\includegraphics[width=0.6\linewidth]{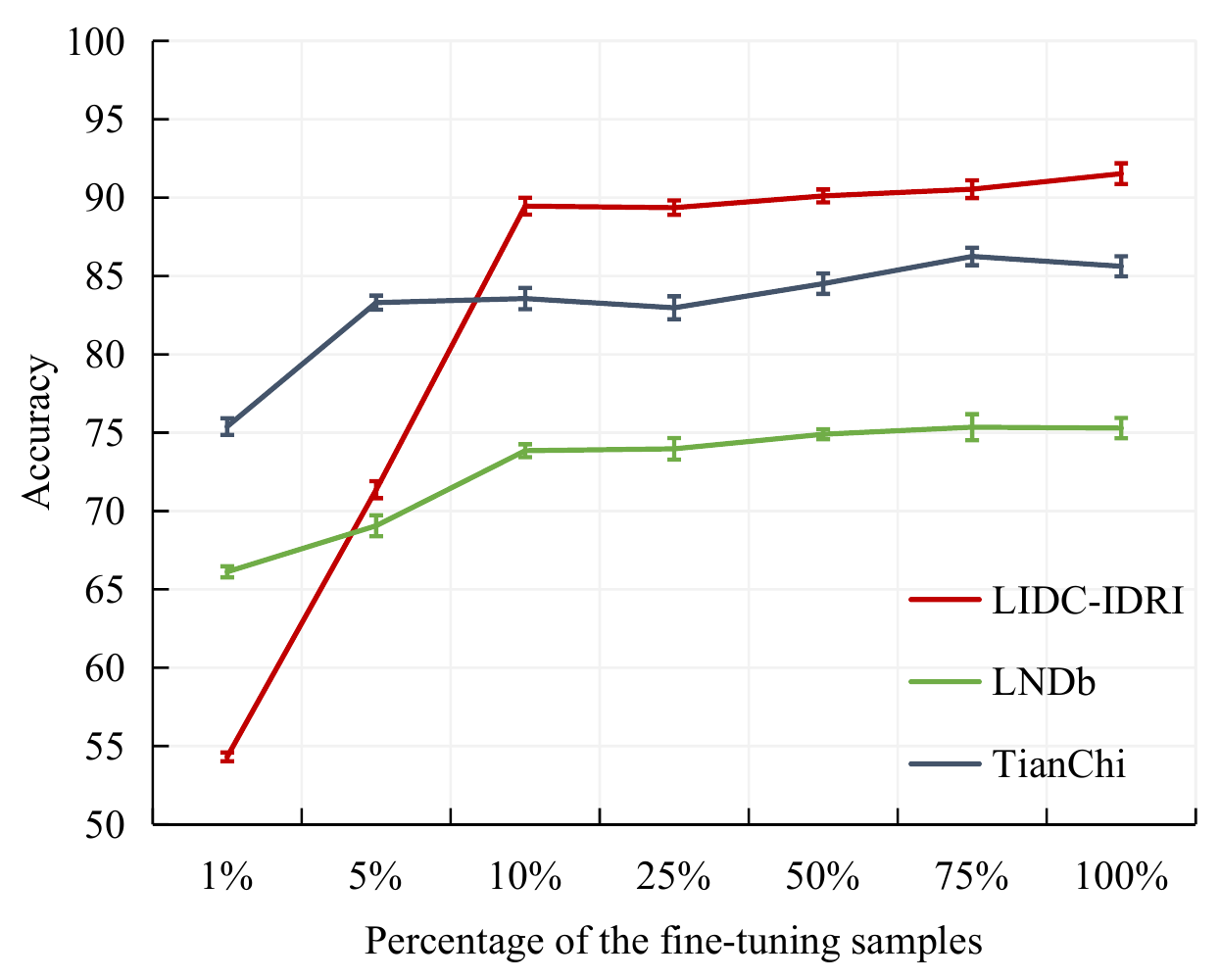}}
\caption{Results of fine-tuning with different numbers of samples.}
\label{mvcnet_fine-tune-appe}
\end{figure}

\subsection{Fine-tuning with Different Percentages of Samples}
We also investigate the performance of MVCNet with respect to the fine-tuning samples on the three datasets and summarize the results in Figure \ref{mvcnet_fine-tune-appe}. We consider 1\%, 5\%, 10\%, 25\%, 50\%, 75\% and 100\% of the fine-tuning samples. According to the overall trend, the accuracy increases as the amount of sample increases. When the sample size is very small (e.g., 1\%), the mean accuracy of MVCNet is poor, especially for LIDC-IDRI (accuracy=54.31\%). The main reason is that there are not enough samples to learn the inter-sample differences. However, the accuracy can reach a rather high point when using 10\% samples.

% \begin{wrapfigure}{l}
% \centerline{\includegraphics[width=\linewidth]{fine-tune_appe.pdf}}
% \caption{Results of fine-tuning with different numbers of samples.}
% \label{mvcnet_fine-tune-appe}
% \end{wrapfigure}

\begin{figure}[h]
\centerline{\includegraphics[width=\linewidth]{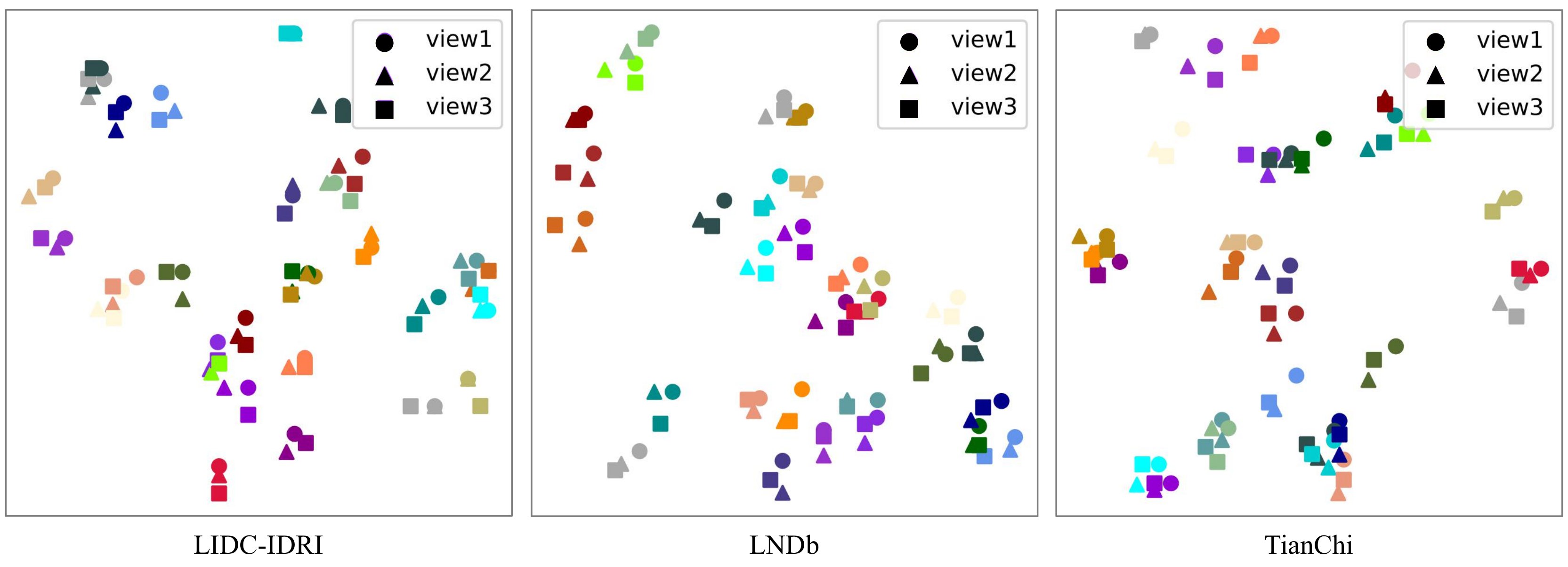}}
\caption{The visualization of the learned representations of the MVCNet on three datasets by t-SNE. We opt for simplicity and show the representations of three views (view1, view2, and view3 in Figure \ref{mvcnet_train}), and randomly sample 30 examples. Each shape represents a view, and different colors represent different lesions. The paired views are fed into MVCNet to obtain the representations, followed by t-SNE to reduce the representation dimension to 2. The closer the distance between the three views of the same lesion, the better the spatial aggregation of the learned representations. Best viewed electronically.}
\label{mvcnet_tsne}
\end{figure}

\subsection{Representation Visualizations}
To visually demonstrate the effect of representation learning, we visualize the representations learned from the three datasets in Figure \ref{mvcnet_tsne}. For a clear demonstration, we randomly sample 30 lesions and three views from each dataset. These views are fed into the network to yield the representations, which are followed by t-SNE \cite{van2008visualizing}. In Figure \ref{mvcnet_tsne}, each shape represents a view, and different colors represent different lesions. The closer the distance between the  views of a same lesion, the better the within-lesion compactness. It is observed from Figure \ref{mvcnet_tsne} that the MVCNet can minimize the distance among views of a same lesion. At the same time, the representations of different lesions do not collapse together. The three datasets all show the same property.

\section{Discussion}
% Include a Discussion that summarizes (but does not merely repeat) your conclusions and elaborates on their implications. There should be a paragraph outlining the limitations of your results and interpretation, as well as a discussion of the steps that need to be taken for the findings to be applied. Please avoid claims of priority.

Learning view-invariant representations for 3D lesions in CT is fundamental in many applications since the local structures of tissues occur at arbitrary views (see Supplementary Figure S2, Figure S3 and Figure S4). MVCNet aggregates multiple 2D views of the same 3D CT scan to make the model make view-invariant predictions. From the perspective of the training scheme, our view-based method is quite different from the transformation-based works. The transformation-based methods can learn useful semantic representations via virtual transformations. However, the transformations are designed empirically and are not universal for all data. For example, color transformation can hurt the fine-grained classification of birds and the rotation is not useful for coarse-grained classification \cite{xiao2020should}. The proposed framework has two advantages. First, it is transformation-free, and we do not spend a lot of time and effort to design and validate the transformations. Inappropriate transformations may decrease the performance of the model. Second, multi-view learning and CT data can be combined logically and smoothly, given that the CT data itself is three-dimensional.
% Learning view-invariant representations for 3D lesions in CT is fundamental in many applications since the local structures of tissues occur at arbitrary views (see Appendix Fig. \ref{mvcnet_nine_views_lidc}, Fig. \ref{mvcnet_nine_views_lndb} and Fig. \ref{mvcnet_nine_views_tianchi}).

Choosing how many views for each batch for training is an important variable, which is associated with the diagnostic accuracy. To understand this, Figure S1 depicts the accuracy on the LIDC-IDRI with respect to the number of views. We find that the diagnostic accuracy also increased as the number of views increased. The result proves our motivation that multiple views can effectively improve the diagnostic performance of 3D lesions. Essentially, the lesion (e.g., lung nodule) is a 3D object and the lesion has different characteristics for different views. Therefore, the challenge in learning 3D lesion representations using multi-view is twofold, namely to maximize the commonality of all views while still preserving the differences between the views. We solve the first challenge by contrast learning, the contrast loss can maximize the similarity between the representations of different views of the same lesion. Modeling for each view can ensure the fairness of the view and learn as many characteristics of each view as possible. Therefore, we can achieve the goal of learning the 3D representations of the lesion by the proposed MVCNet.
% To understand this, Fig. \ref{mvcnet_views} depicts the accuracy on the LIDC-IDRI with respect to the number of views.

The proposed MVCNet is a novel multi-view contrastive learning framework for learning representations from volumetric CT data. Unlike existing SSL methods, MVCNet works in a transformation-free manner and no explicit pretext task is needed. Our approach is evaluated on three CT datasets for disease diagnosis. Experimental results show that our method outperforms state-of-the-art SSL methods, indicating the superiority of MVCNet in \emph{unsupervised representation learning}. Being fine-tuned with a small percentage of the datasets (10\%), our model is comparable to the 3D fully supervised model, demonstrating its superiority in \emph{small-data scenarios} and the potential of reducing the annotation efforts in 3D CAD systems.

The main insufficiency of our work is that MVCNet is designed for the lesion identification task and is not optimized for the localization task on the whole CT volumes. Lesion identification is similar to object recognition in computer vision in that both are prospect dominant. The effectiveness of multi-view methods in lesion localization (often with background dominance) needs to be explored. The future work is twofold. First, in theory, we can expand MVCNet to any number of views from unlimited orientations. We will investigate the number of views and the correlations between views. Second, we will adapt MVCNet to more tasks (lesion localization and segmentation) and evaluate it on more imaging modalities such as the magnetic resonance imaging.

\section{Materials and Methods}
% The materials and methods section should provide sufficient information to allow replication of the results. This section should be broken up by subheadings. Under exceptional circumstances, when a particularly lengthy description is required, a portion of the materials and methods can be included in the Supplementary Materials. 

\subsection{LIDC-IDRI}
The LIDC-IDRI \footnote{https://wiki.cancerimagingarchive.net/display/Public/LIDC-IDRI} is a commonly-used CT dataset. It contains 1,018 cases used for lung nodules diagnosis \cite{armato2011lung}. The malignancy of each nodule was evaluated by up to four experienced radiologists in two stages with a 5-point scale from benign to malignant. All suspicious lesions are categorized to three types according to the diameter in long axis: non-nodule $ >= 3 mm $, nodule $ < 3 mm $, and nodule $ >= 3 mm $. We only consider nodules $ >= 3 mm $ in diameter, since nodules $ < 3 mm $ were not considered to be clinically relevant. The slice thickness of CT scans ranges from $ 0.6 mm $ to $ 5.0 mm $ with a median of $ 2.0 mm $. In this study, scans with slice thickness thicker than $ 2.5 mm $ were eliminated, as this could easily lead to the omission of small nodules \cite{setio2016pulmonary, naidich2013recommendations}. Following the procedures used in previous studies \cite{xie2018knowledge}, we calculate the mean malignancy degree ($ d $) of a nodule which was annotated by at least three radiologists, and annotate a nodule whose $ d < 3 $ as benign, a nodule whose $ d = 3 $ as uncertain and a nodule whose $ d > 3 $ as malignant. There are totally 369 benign, 405 uncertain and 335 malignant nodules. To reduce the impact of uncertain evaluation, we exclude all uncertain lung nodules from the dataset.
% In the initial blind reading phase, each radiologist independently reviews each CT scan and marks lesions that fall into one of three categories (during the labeling process, the results are categorized into three categories: nodules $ >= 3 mm $, nodules $ < 3 mm $, and non-nodules $ >= 3 mm $). In the second stage, each radiologist independently reviews his or her own labeling as well as anonymous labeling by three other radiologists to provide a final opinion. The goal of this process was to identify all pulmonary nodules in each CT scan as completely as possible, without forcing a consensus ~\cite{}.

\subsection{LNDb}
The LNDb \footnote{https://lndb.grand-challenge.org/Data/} dataset contains a total of 294 CT scans \cite{pedrosa2019lndb}. The annotation method is adapted from the LIDC-IDRI. Each CT scan is read by at least one radiologist to evaluate the nodules with a 5-point scale from benign to malignant. Therefore, we process the LNDb dataset in the LIDC-IDRI way. There are few nodules annotated by at least three radiologists. Therefore, we calculate the mean malignancy degree ($ d $) of all annotated nodules. Finally, there are 451 benign and 768 malignant nodules after removing the uncertain nodules.

\subsection{TianChi}
The TianChi lung multi-disease diagnosis dataset \footnote{https://tianchi.aliyun.com/competition/entrance/231724/information?lang=en-us} contains a total of 1,470 CT scans with four diseases. In the annotation file, radiologists record the center coordinates, size (consisting of three diameters) and disease category. The annotation process is divided into two stages: two physicians performed the original annotation, then a third independent physician performed the disambiguation to ensure the consistency of the data annotation. We use all the annotated lesions in the dataset. There are 3,264 nodules, 3,613 Streak shadows, 4,201 arteriosclerosis or calcification and 1,140 lymph node calcification.

\begin{figure}[h]
\centerline{\includegraphics[width=\linewidth]{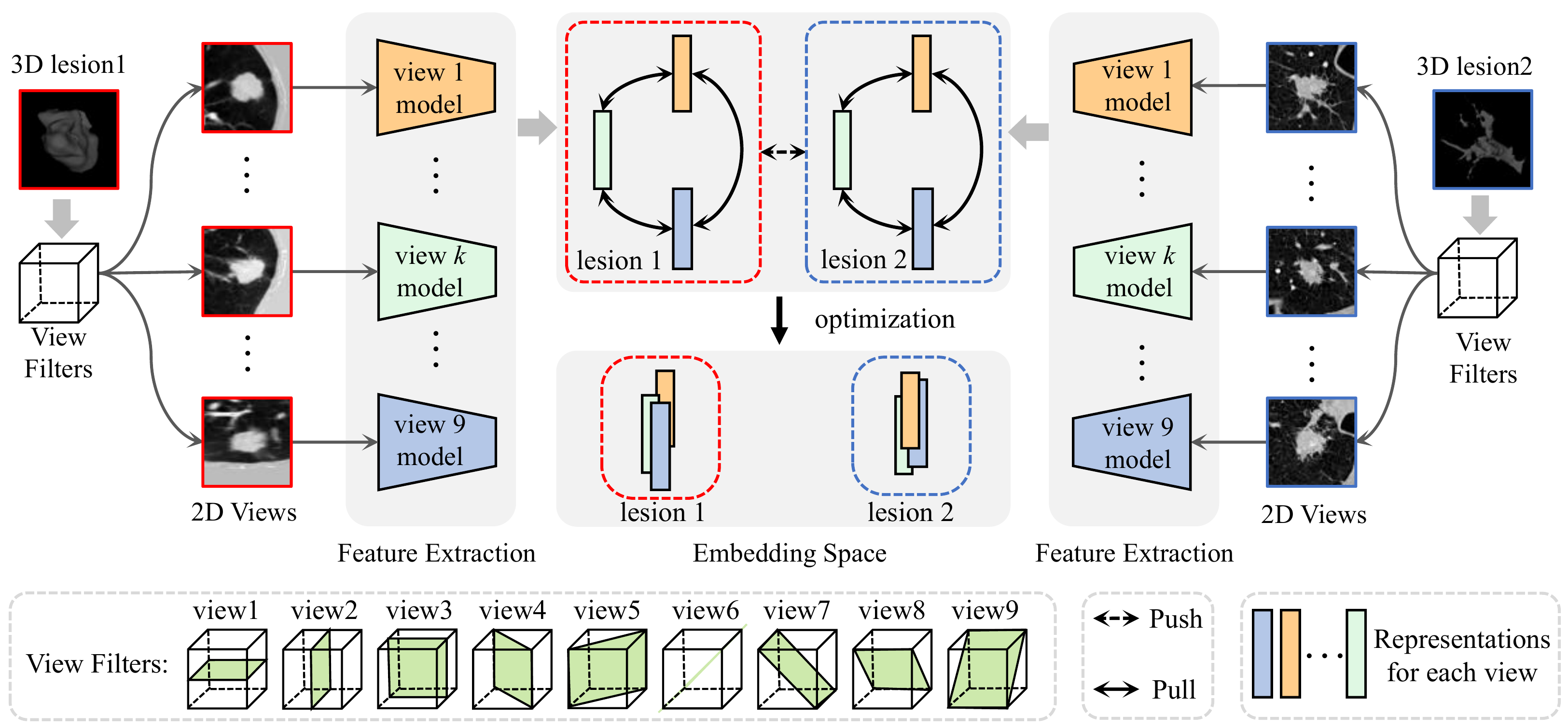}}
\caption{Illustration of the training stage of the MVCNet. For simplicity, we show two 3D lesions. They are fed into a set of view filters with nine orientations to generate multiple 2D views, respectively. Inherited from \cite{tian2020contrastive}, each view has a private encoder and a projector to generate representations in a embedding space. We make views of the same lesions attract each other and views of different lesions repeal each other. Being optimized by a contrastive loss, the shared embedding space is endowed with well local-aggregating properties and the spread-out properties of representations are preserved. Note that each view model is shared across lesions.}
\label{mvcnet_train}
\end{figure}

\subsection{Implementation Details}
We extract 2D views of lesion volumes from different orientations, and resize each view to $ 224 \times 224 $. We use a tiny AlexNet \cite{krizhevsky2012imagenet} as our base encoder without fully connected layers. The model contains three convolutional layers, and the number of channels produced by the convolution is 48, 192, 128, respectively. For the projector, we use a light neural network that maps representations to the embedding space where contrastive loss is applied. The projector includes three fully connected layer, and the first two are followed by a batch normalization layer and activated by ReLU \cite{nair2010Rectified}. The output dimensions of the layers are 2,048, 2,048 and 128. After the projector, we apply a $ {\ell}_2 $-normalize. Stochastic gradient descent with a base learning rate of 0.1 is used to optimize the encoder and projector. The training epoch is set to 240, and when reaching 120, 160, and 200 epochs, the learning rate decays by a rate of 0.1. We use a batch size of 64, and the weight decay rate is $ 1e-4 $. The temperature $ \tau $ is set to 0.07 by following previous works \cite{tian2020contrastive, wu2018unsupervised}. Most of the details are based on the Contrastive Multiview coding (CMC) implementation \cite{tian2020contrastive}. All the experiments are conducted with PyTorch\footnote{https://pytorch.org/} using a single Tesla V100 32GB GPU.
%  The other parameters correspond to the first, third and fifth convolutional layers of Alexnet, respectively.

% \begin{figure}[h]
%     \centering
%     \caption{Illustration of the training stage of the MVCNet. For simplicity, we show two 3D lesions. They are fed into a set of view filters with nine orientations to generate multiple 2D views, respectively. Inherited from \cite{tian2020contrastive}, each view has a private encoder and a projector to generate representations in a embedding space. We make views of the same lesions attract each other and views of different lesions repeal each other. Being optimized by a contrastive loss, the shared embedding space is endowed with well local-aggregating properties and the spread-out properties of representations are preserved. Note that each view model is shared across lesions.}
%     \label{fig:2}
% \end{figure}

We show the flowchart of the proposed MVCNet in Figure \ref{mvcnet_train}. It is notable that we focus on the lesion diagnosis task and assume that the detection of suspected lesions has been completed. Nine views are extracted from each lesion volume from different orientations. Next, we construct a convolutional neural network to learn representations by minimizing a contrastive loss. Finally, we evaluate the representations with 1) linear evaluation by training a classification head upon the MVCNet with fixed parameters and 2) fine-tuning the model with a small fraction of annonated data.
% Then, we construct the MVCNet model to learn potential representation by pulling the representation distance of these views extracted from a same volume and pushing the representation distance of views extracted from different volumes.
% The pseudo-code is shown in Algorithm. \ref{mvcnet_code}.

\subsection{View Extraction}
Before extracting lesion volumes, we process the datasets based on a same procedure since all the datasets are for lung CT scans. Following previous studies, we truncate the range of Hounsfield (HU) values to $ [-1000, 400] $ and then scaled it to $ [0, 1] $ to reduce the influence of other organs \cite{setio2016pulmonary, xie2018knowledge}. Then, we resize the pixel resolution of all CT scans to $ 1 mm \times 1 mm \times 1 mm $, which corresponds to the most common resolution of CT scans \cite{xie2018knowledge, shen2017multi}.

After preprocessing, we extract the lesion volumes according to the lesion diameter. The LIDC-IDRI and LNDb are used to diagnose lung nodules. We set the size of each lesion volume to $ 64 mm \times 64 mm \times 64 mm $ since the diameter of nodules is generally between $ 3 mm $ and $ 30 mm $ \cite{xie2018knowledge}. For the TianChi dataset, there are four different types of diseases, and we calculate the lesion size based on the longest diameter of each lesion. Assuming that the longest diameter of the lesion is $ d mm $, the size of the lesion volume is $ (d + 20) mm \times (d + 20) mm \times (d + 20) mm $. Finally, to avoid the empirical design of transformations and retain the 3D characteristics of lesions, we introduce a new approach to extract nine views of a lesion volume from different orientations \cite{setio2016pulmonary}. We show the nine orientations as view filters at the bottom of Figure \ref{mvcnet_train}. Supplementary Figure S2, Figure S3 and Figure S4 present some examples of 2D views for 3D lesions.
% Supplementary Figure \ref{mvcnet_nine_views_lidc}, Fig. \ref{mvcnet_nine_views_lndb} and Fig. \ref{mvcnet_nine_views_tianchi} present some examples of 2D views for 3D lesions.

\subsection{Contrastive Loss}
Contrastive learning aims at constructing a latent embedding space for separating samples from different clusters in an unsupervised manner. Like most of the existing works \cite{tian2020contrastive, chen2020simple, he2020momentum}, we use a contrastive loss to enhance the similarity within lesions and separability between lesions.

% Unlike most current contrastive learning methods \cite{b19, b20}, w
% Following the CMC \cite{tian2020contrastive}, we build an independent model for each view instead of sharing parameters for all views.
Following the CMC \cite{tian2020contrastive}, we build a multi-view contrastive learning model to learn 3D representation through 2D views. In this way, we can maximize the mutual information and preserve the differences among the views. We consider the 3D characteristics of lesions by combining the representations of all the 2D views for downstream tasks.
% , where $ n $ represent the number of views used in the framework, $ n \in [1, 9] $.

% Each model in the MVCNet is composed by an encoder $ f(\cdot) $ and a projector $ g(\cdot) $.
The model is composed by an encoder $ f(\cdot) $ and a projector $ g(\cdot) $. We illustrate the process with two views (represented by $ v_{1} $ and $ v_{2} $, e.g., view1 and view2 in Figure \ref{mvcnet_train}). Following the previous works \cite{chen2020simple, tian2020contrastive}, we learn the representation vector $ y $ by the encoder from each view. i.e., $ y_{1} = f_{1}(v_{1}) $ and $ y_{2} = f_{2}(v_{2}) $. Then, we map the representation to a projection $ z $ by the projector, i.e., $ z_{1} = g_{1}(y_{1}) = g_{1}(f_{1}(v_{1})) $ and $ z_{2} = g_{2}(y_{2}) = g_{2}(f_{2}(v_{2})) $. We adopt a tiny AlexNet \cite{krizhevsky2012imagenet} as our encoder, and the $ y \in \mathbb{R}^{d} $ is the output after the last convolutional layer. The projector is a multi-layer perception (MLP) with three hidden layers and a $ \ell_{2} $-normalization. The MLP can be seen as the fully connected layers of AlexNet. The model is initialized randomly. Same as most SSL methods \cite{chen2020simple, he2020momentum, grill2020bootstrap}, the role of the projector is to eliminate the semantically irrelevant low-level information in the representation. After pretraining, the projector is discarded and the representations of the encoder are used for downstream tasks.

We randomly sample a batch of $ N $ lesions and define the contrastive task within the batch, and so the number of views is $ 2 N $ in each batch. We denote a given batch as $ \left\{v_{1}^{i}, v_{2}^{i}\right\}_{i=1}^{N} $. Then, the views of the same lesion form a positive pair (e.g., $ \left\{v_{1}^{1}, v_{2}^{1}\right\} $), and we treat the left $ 2 (N - 1) $ views of different lesions as negative samples, which is in consistent with the previous studies \cite{tian2020contrastive, chen2020simple}.

We exploit a contrastive loss to achieve the high similarity for positive pairs and low similarity for negative pairs. We calculate the cosine similarity ($ sim(\cdot) $) \cite{tian2020contrastive} as the metric to evaluate the similarity of views. Assuming $ \operatorname{sim}(\boldsymbol{u}, \boldsymbol{v})=\boldsymbol{u}^{\top} \boldsymbol{v} /\|\boldsymbol{u}\|\|\boldsymbol{v}\| $, $ \boldsymbol{u} $ and $ \boldsymbol{v} $ are the two different $ \ell_{2} $ normalized views, we can define the objective function for a positive pair $ \{v_{1}^{i}, v_{2}^{i}\} $ as
\begin{equation}
    \mathcal{L}_{(v_{1}^{i}, v_{2}^{i})} = -\log \frac{\exp \left(\operatorname{sim}\left(\boldsymbol{z}_{1}^{i}, \boldsymbol{z}_{2}^{i}\right) / \tau\right)}{\sum_{k=1}^{N} \mathds{1}_{[k \neq i]} \exp \left(\operatorname{sim}\left(\boldsymbol{z}_{1}^{i}, \boldsymbol{z}_{2}^{k}\right) / \tau\right)} ,
    \label{loss_positive_pair}
\end{equation}
where $ \tau $ denotes a temperature parameter to adjust the dynamic range of the loss. $ \mathds{1}_{[k \neq i]} $ is an indicator function to determine whether $ v_{1}^{i} $ and $ v_{2}^{i} $ belong to a same lesion with $ \mathds{1}_{[k \neq i]} \in \{0, 1\} $. If $ v_{1}^{i} $ and $ v_{2}^{i} $ are extracted from the same lesion, $ \mathds{1}_{[k \neq i]} = 1 $, and otherwise $ \mathds{1}_{[k \neq i]} = 0 $. The final loss is computed across all positive pairs in a batch, and $ \mathcal{L}_{(v_{1}^{i}, v_{2}^{i})} $ treats the $ v_{1}^{i} $ as anchor and enumerates over $ v_{2}^{i} $. Symmetrically, we can get $ \mathcal{L}_{(v_{2}^{i}, v_{1}^{i})} $ by anchoring $ v_{2}^{i} $. Therefore, the final loss is denoted by adding the two losses up:
\begin{equation}
    \mathcal{L}_{v^{i}} = \mathcal{L}_{(v_{1}^{i}, v_{2}^{i})} + \mathcal{L}_{(v_{2}^{i}, v_{1}^{i})} ,
    \label{loss_vi_add_vj}
\end{equation}
where $ v^{i} $ represents all views extracted from the same lesion, and $ v^{i} = \{v_{1}^{i}, v_{2}^{i}\}$.

% In total, we decompose nine views from a lesion volume as shown in Fig. \ref{mvcnet_train}(b).
MVCNet is a general framework that can be applied to different numbers of views. In this paper, we limit the maximum view number to nine. Suppose we have a collection of $ M ( 3 \leq M \leq 9) $ views in a batch, $ V = \{v_{1},..., v_{m}, ..., v_{M}\} $. We treat the view that we want to optimize as the anchor, e.g., $ v_{m} $. After that, we form the positive pairs between anchor $ v_{m} $ and each other view $ v_{k, (k \neq m)} $. According to Equation \ref{loss_positive_pair}, we compute the loss by summing up all the positive pairs:
\begin{equation}
    \mathcal{L}_{\left(v_{m}^{i}, v^{i}\right)} = \sum_{j=1, j \neq m}^{M} \mathcal{L}\left(v_{m}^{i}, v_{j}^{i}\right) ,
    \label{loss_vi_over_all_views}
\end{equation}
where $ \mathcal{L}_{\left(v_{m}^{i}, v^{i}\right)} $ is the loss between the anchor $ v_{m}^{i} $ and other views. Considering all views act as anchor in turn, the objective function $ \mathcal{L}_{V} $ is formulated as:
\begin{equation}
    \begin{aligned}
        \mathcal{L}_{V} &= \sum_{m = 1}^{M} \mathcal{L}_{\left(v_{m}^{i}, v^{i}\right)} \\
        % &= \sum_{m = 1}^{M} \sum_{j = 1, j \neq m}^{M} \mathcal{L}(v_{m}^{i}, v_{j}^{i}) \\
        &= \sum_{1 \leq m, j \leq M, m \neq j} \mathcal{L}(v_{m}^{i}, v_{j}^{i}) ,
        % \mathcal{L} = \sum_{1 \leq m \leq M} \mathcal{L}_{v_{k}} = \sum_{1 \leq i<j \leq V} \mathcal{L}\left(v_{i}, v_{j}\right) ,
    \end{aligned}
    \label{loss_final}
\end{equation}
where $ \mathcal{L}_{V} $ is the final loss when we only consider the $ i $-th lesion in the batch. Taking a seep forward, we formulate the objective function in the batch as:
\begin{equation}
        \mathcal{L} = \frac{1}{2 N} \sum_{n = 1}^{N} \mathcal{L}_{V}. \\
    \label{loss}
\end{equation}
We optimize all the encoders and projectors with the objective function $ \mathcal{L} $. From Equation \ref{loss} . Empirically, the more views we consider, the heavier the computational cost.

\begin{figure}[h]
\centerline{\includegraphics[width=0.7\linewidth]{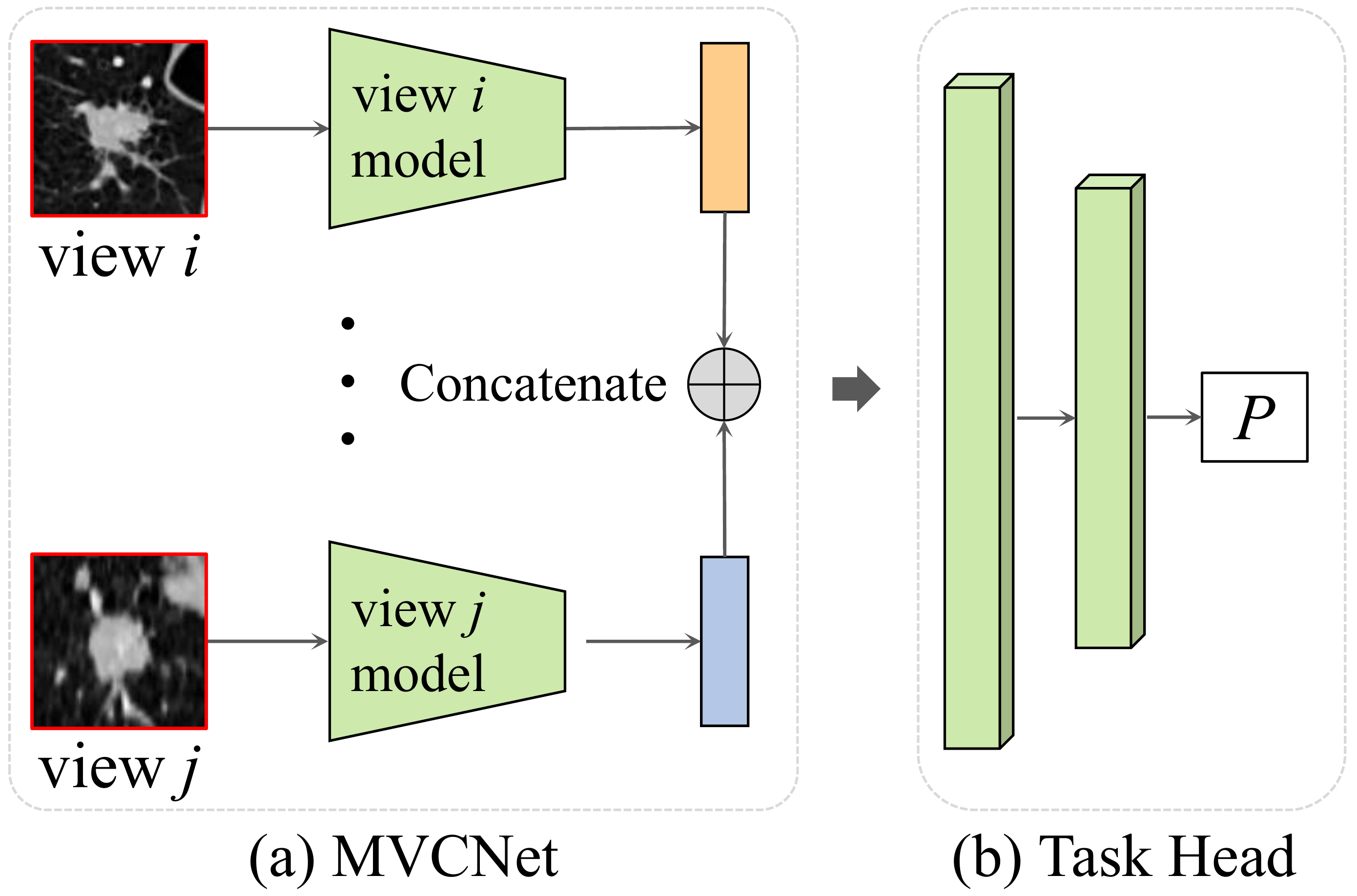}}
\caption{The inference stage of MVCNet. We extract representations of each view and concatenate them together. Then we feed the concatenated representation to a simple task head, computing the disease probability with a softmax function.}
\label{mvcnet_inference}
\end{figure}

\subsection{Target Task}
The goal of the target task is to evaluate the quality of the representations learned by the MVCNet. Inspired by \cite{tian2020contrastive, chen2020simple}, we evaluate the representation using the diagnostic accuracy by adding a simple classification head (see Figure \ref{mvcnet_inference}). It only contains a fully connected layer trained from scratch. The input of the head is the concatenated representations from different views. It is noteworthy that we extract representations from the encoder $ f(\cdot) $ rather than the projector $ g(\cdot) $. After pretraining, the projector is discarded. Then, we feed the concatenated representations to the head. The dimension of the input representation is subject to the number of views. Namely, the input dimension is $ N \times d $, where $ N $ represents the number of views, and the dimension from each view is $ d $. Finally, we get the disease probability with a softmax function.

\section*{Conflicts of Interest}
The authors declare no competing interests.

\section*{Acknowledgments}
This work was supported by the National Natural Science Foundation of China (62020106015) ,the Strategic Priority Research Program of CAS (XDB32040000), the Zhejiang Provincial Natural Science Foundation of China (LQ20F030013), and the Ningbo Public Service Technology Foundation, China (202002N3181), Ningbo Natural Science Foundation, China (202003N4270). The authors appreciate the doctors at department of radiology, HwaMei Hospital, University of Chinese Academy of Sciences for their comments on learning representations with limited annotated data.

\subsection*{Author Contributions} 
J. li conceived the idea and designed the experiments. P. Zhai conducted the experiments. H. Cong, G. Zhao, C. Fang, T. Cai and H. He contributed equally to this work.

\section*{Supplementary Materials}
Table S1: Several typical self-supervised learning approaches proposed to process natural and medical images. Table S2: The augmentations of natural images and medical imaging. Table S3-S4: Comparison of MVCNet with state-of-the-art SSL methods in LNDb and TianChi. Figure S1: The accuracy with respect to view numbers on LIDC-IDRI. Figure S2-S4: Visualization of nine views of malignant and benign nodules on LIDC-IDRI, LNDb and TianChi.

\printbibliography

\end{document}